\ifdefined\pdfminorversion
\fi
\documentclass[letterpaper]{article}
\usepackage{spconf,amsmath,amssymb,graphicx,booktabs,microtype,cite,xspace}
\usepackage[T1]{fontenc}
\usepackage{times,newtxmath}
\usepackage[hidelinks]{hyperref}
\usepackage{etoolbox,flafter,float}
\usepackage[spread,noshrink]{flushend}
\microtypesetup{protrusion=false}
\apptocmd{\thebibliography}{\setlength{\itemsep}{0pt}\setlength{\parsep}{0pt}}{}{}
\graphicspath{{figures_reviewed/}{figures/}{./}}
\newcommand{\paperfigure}[3]{%
  \IfFileExists{#3}{\includegraphics[width=#1]{#3}}{%
    \IfFileExists{figures/#3}{\includegraphics[width=#1]{figures/#3}}{%
      \includegraphics[width=#1]{#2}}}}

\newcommand{\figOneScale}{1.00}   
\newcommand{\figTwoScale}{0.90}   
\newcommand{\figThreeScale}{0.95} 
\newcommand{\figFourScale}{1}  
\newcommand{\figFiveScale}{0.95}  
\newcommand{\tableScale}{0.96}   

\newcommand{\method}{SwitchPFN\xspace}
\newcommand{\R}{\mathbb R}
\newcommand{\D}{\mathcal D}
\newcommand{\mat}[1]{\mathbf{#1}}
\newcommand{\vect}[1]{\boldsymbol{#1}}
\AtBeginEnvironment{equation}{%
  \setlength{\abovedisplayskip}{4pt plus 1pt minus 1pt}%
  \setlength{\belowdisplayskip}{3pt plus 1pt minus 1pt}%
  \setlength{\abovedisplayshortskip}{1pt plus 1pt minus 1pt}%
  \setlength{\belowdisplayshortskip}{3pt plus 1pt minus 1pt}}

\newcommand{\unnumsection}[1]{\par\vspace{0pt}{\centering\normalsize\bfseries\MakeUppercase{#1}\par}\vspace{0pt}\noindent\ignorespaces}
\title{SwitchPFN: Shared Switching Dynamics for Frozen In-Context Time Series Classification}
\name{\fontsize{11.5}{13}\selectfont\begin{tabular}{@{}c@{}}
Zhenyi Zhu$^{1,*}$, Jacqueline Pang$^{2,*}$, Peilin Shen$^{3,*}$, Tianyi Song$^{4,\dagger}$, Tingwei Zhang$^5$,\\
Keyi Hu$^6$, Kangjun Yin$^7$, Shiwei Pu$^8$, Yingbo Zhou$^9$, Chen Shao$^{10,\dagger}$
\end{tabular}
\thanks{\raggedright$^*$Equal contribution. \smash{$^\dagger$}Corresponding authors: \href{mailto:tianyi.song.24@ucl.ac.uk}{tianyi.song.24@ucl.ac.uk}; \href{mailto:chen.shao2@kit.edu}{chen.shao2@kit.edu}.}}
\address{
\fontsize{10.5}{13}\selectfont \mbox{$^1$The Chinese University of Hong Kong, Hong Kong}\quad \mbox{$^2$Cornell University, USA}\quad \mbox{$^3$Southeast University, China}\\
\fontsize{10.5}{13}\selectfont \mbox{$^4$University College London, London, UK}\quad \mbox{$^5$Dalian Polytechnic University, China}\quad \mbox{$^6$Hainan Normal University, China}\\
\fontsize{10.5}{13}\selectfont \mbox{$^7$Anhui Agricultural University, China}\quad \mbox{$^8$Beijing University of Posts and Telecommunications, China}\\
\fontsize{10.5}{13}\selectfont \mbox{$^9$East China Normal University, China}\quad \mbox{$^{10}$Karlsruhe Institute of Technology, Germany}}
\makeatletter
\newcommand{\methodsection}{\@startsection{section}{1}{\z@}%
  {-8pt plus 1pt minus 1pt}{5pt plus 1pt minus 1pt}{\normalfont\normalsize\bfseries}}
\newcommand{\methodsubsection}{\@startsection{subsection}{2}{\z@}%
  {-7pt plus 1pt minus 1pt}{4pt plus 1pt minus 1pt}{\normalfont\normalsize\bfseries}}
\newlength{\figureCaptionGap}
\newcommand{\tighttablecaption}{\patchcmd{\@makecaption}{\vskip 10pt}{\vskip 3pt}{}{\PackageError{SwitchPFN}{Table caption spacing patch failed}{}}}
\newcommand{\tightfigurecaption}{%
  \long\def\@makecaption##1##2{%
    \vskip\figureCaptionGap
    \setbox\@tempboxa\hbox{##1. ##2}%
    \nointerlineskip
    \ifdim\wd\@tempboxa>\hsize
      \noindent ##1. ##2\par
    \else
      \hbox to\hsize{\hfil\box\@tempboxa\hfil}%
    \fi}}
\patchcmd{\@maketitle}{\large \bf \@title}{\fontsize{14}{16}\selectfont\bfseries\@title}{}{\PackageError{SwitchPFN}{Title patch failed}{}}
\patchcmd{\@maketitle}{\vskip 2em}{\vskip 2.55em}{}{\PackageError{SwitchPFN}{Title position patch failed}{}}
\patchcmd{\@maketitle}{\vskip 1.5em}{\vskip 3pt}{}{\PackageError{SwitchPFN}{Title-to-author spacing patch failed}{}}
\patchcmd{\@maketitle}{\begin{tabular}[t]{c}}{\begin{tabular}[t]{@{}c@{}}}{}{\PackageError{SwitchPFN}{Title table padding patch failed}{}}
\patchcmd{\@maketitle}{\@name \\ \@address}{\@name \\[\dimexpr-0.5em-1.72pt\relax] \@address}{}{\PackageError{SwitchPFN}{Author-to-affiliation spacing patch failed}{}}
\patchcmd{\@maketitle}{\vskip 1.5em}{\vskip 0.6696pt}{}{\PackageError{SwitchPFN}{Frontmatter-to-body spacing patch failed}{}}
\makeatother
\begin{document}
\ninept
\ifdefined\XeTeXversion
  
\fi
\maketitle
\flushbottom
\linespread{.95}\selectfont
\nocite{Hannun2019CardiologistLevelAD,7961240,c64eb143d0dc44f0a027e73b13a8a486,inceptiontime,timesnet,tst,ts2vec,moment,pfn2022,nagler2023,tabpfn2023,tabpfn2025,tabpfn3,ts2tabpfn,rocketpfn,rocket,minirocket,multirocket,hydra,Shao_2025,shao2026gnnsfailquantifyingovercoming,dmd,koopman,pca,ridge,dtw,xgboost,lstm,informer,dlinear,signature,deepsignature,uea2018,ucr2019,li2023time}

\begin{abstract}
Tabular foundation models (TFMs) provide a promising route to time-series classification, but their effectiveness depends on how sequential data are converted into tabular representations. Existing representations face two challenges: global aggregation can lose the order of temporal evolution, while features computed in independently fitted coordinate systems may not have consistent meanings across sequences. We therefore view representation design for TFMs as a problem in its own right: the representation should preserve local temporal transitions while maintaining a shared feature definition across samples. We propose \method, which learns a shared projection and regime codebook from the training sequences, making local dynamic operators and transition features directly comparable across samples. Across the evaluated benchmarks, \method achieves the highest mean accuracy among the evaluated methods, improving over the strongest baseline by 4.47\% relatively. Ablation studies, parameter sensitivity analyses, and reduced-training-data experiments further examine the contributions of the representation, its main design choices, and its behavior when labeled data are limited.
\end{abstract}


\vspace{-6pt}
\section{Introduction}
\vspace{-3pt}
Classifying temporal signals is a crucial task across science and engineering, with applications in ECG arrhythmia detection \cite{Hannun2019CardiologistLevelAD}, sleep staging \cite{7961240}, and astronomical transient classification \cite{c64eb143d0dc44f0a027e73b13a8a486}. It assigns a label to a sequence from its signal values and temporal changes. InceptionTime uses convolutional patterns and TimesNet uses periodic variations~\cite{inceptiontime,timesnet}. Transformer and contrastive pretraining provide reusable temporal representations~\cite{tst,ts2vec,moment}.

A recent line of work in time series classification separates what a model sees from how it predicts, pairing a temporal representation with a pretrained, general-purpose predictor. The predictor side builds on prior-data fitted networks (PFNs), which are pretrained on synthetic tasks to perform prediction in context, approximating Bayesian inference without task-specific training~\cite{pfn2022,nagler2023}. TabPFN brings this principle to tabular data: given labeled rows as context, a frozen transformer classifies new rows in a single forward pass~\cite{tabpfn2023,tabpfn2025,tabpfn3}. TS2TabPFN and RocketPFN use such TFMs for time series by first converting each series into a table row~\cite{ts2tabpfn,rocketpfn}. ROCKET, MiniRocket, MultiRocket, and HYDRA summarize a series through the responses of random or structured convolutional kernels~\cite{rocket,minirocket,multirocket,hydra}. These combinations achieve strong performance, motivating a closer look at how their representations encode changes in local evolution. Changing temporal dependencies also challenge forecasting models~\cite{Shao_2025,shao2026gnnsfailquantifyingovercoming}. This leads to the research question we address in this work:
\par
\begingroup
\linespread{1.03275}\selectfont
\setlength{\parskip}{0pt}
\vspace{2pt}
{\centering\itshape How can local evolution in a time series, and changes in that evolution, be effectively encoded for a frozen tabular foundation model?\par}
\vspace{2pt}\noindent\ignorespaces
To answer this question, the representation must capture both how local patterns unfold over time and properties of their dynamics that remain comparable across series. For a TFM, this means constructing columns with the same meaning in every row. Two challenges follow: 1) How often local patterns occur does not determine the order in which they occur [Fig.~\ref{fig:motivation}(a)];  2) Any representation whose coordinates are fitted per series, for example by PCA or DMD, can produce operator features that are not comparable across rows, since operator coefficients depend on the coordinates in which the state is expressed~\cite{dmd,koopman}. For an invertible coordinate change $\mat{Q}$ with $\vect{z}_t=\mat{Q}\vect{z}'_t$, \vspace{-5pt}
\begin{equation}
 \setlength{\abovedisplayskip}{6pt}
 \setlength{\belowdisplayskip}{6pt}
 \setlength{\abovedisplayshortskip}{6pt}
 \setlength{\belowdisplayshortskip}{6pt}
 \vect{z}_{t+1}=\mat{A}\vect{z}_t\quad\Longrightarrow\quad
 \vect{z}'_{t+1}=\mat{Q}^{-1}\mat{A}\mat{Q}\vect{z}'_t,
 \label{eq:basis}
\end{equation}
so identical dynamics can yield different matrix entries [Fig.~\ref{fig:motivation}(b)]. Discrete states face the same problem, since a codebook fitted per series is defined only up to a relabeling of its states. We therefore fit the projection and the regime codebook once on all training series and share them across every series, giving each feature a common meaning [Fig.~\ref{fig:motivation}(c)].
\par
\vspace{5pt}
\setlength{\intextsep}{0pt}
\enlargethispage{2pt}
\makeatletter
\patchcmd{\float@endH}{\box\@currbox\vskip\intextsep}
  {\box\@currbox\nobreak\vskip\intextsep}{}
  {\PackageError{SwitchPFN}{First figure pagination patch failed}{}}
\makeatother
\begin{figure}[H]
 \tightfigurecaption
 \centering\paperfigure{\figOneScale\columnwidth}{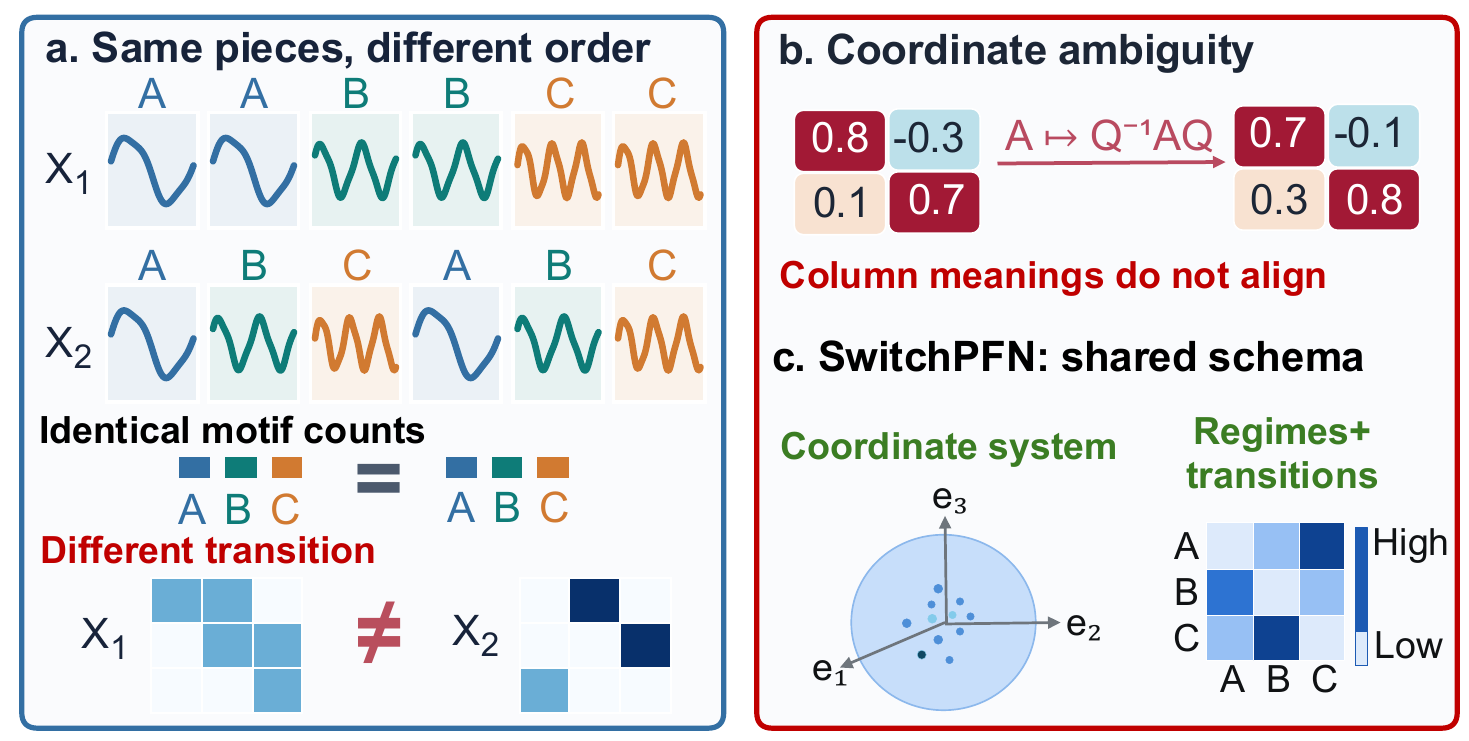}{motivation(6).pdf}\par
 \caption{Two challenges in representing time series in TFM. (a) Aggregating local patterns can preserve their frequency while losing the order in which they occur. (b) Fitting a separate coordinate system for each sequence makes operator features ambiguous across rows. (c) Proposed Designs: \method with a shared projection and regime codebook preserve comparable feature meanings across sequences.}
 \label{fig:motivation}
\end{figure}
\endgroup
\begin{figure*}[!t]
 \tightfigurecaption
 \centering\paperfigure{\figTwoScale\textwidth}{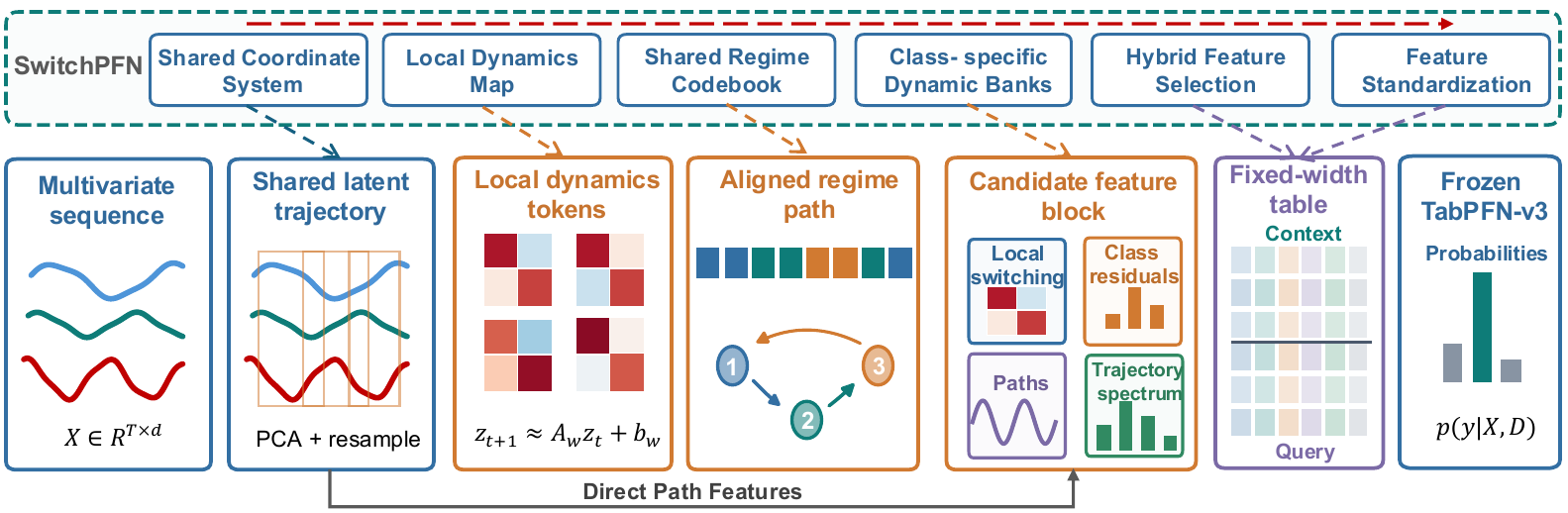}{framework(7).pdf}\par
 \caption{Overview of the \method framework. \method maps each sequence to a fixed-width table for frozen TabPFN. The top row contains training-fitted objects; the bottom row follows the sequence through shared coordinates, local dynamics, regimes, and four feature groups. The direct path branch retains ordered trajectory information.}
 \label{fig:framework}
\end{figure*}
To address these issues, we propose \method, a compact representation for in-context time series classification [Fig.~\ref{fig:motivation}(c)]. It describes the order in which local patterns occur and how the local dynamics switch between regimes, using a shared projection and a shared regime codebook. Our contributions are threefold: 
\begin{list}{\textbullet}{\setlength{\leftmargin}{1.1em}\setlength{\labelsep}{0.45em}\setlength{\topsep}{2pt}\setlength{\partopsep}{0pt}\setlength{\itemsep}{1pt}\setlength{\parsep}{0pt}\interlinepenalty=10000}
 \item We formulate representation design for tabular foundation models as a distinct problem. We identify two complementary requirements: features must capture the temporal order of local patterns and each column must carry the same meaning across sequences.
 \item We develop \method to combine switching dynamics, class residuals, ordered paths, and signal summaries in a fixed-budget table for TabPFN. Class residuals use a \textbf{leave-one-sequence-out} reference for training rows.
\item On eight UEA datasets, \method achieves 82.63\% mean test accuracy, 3.54 percentage points higher than the second-best model. Ablation studies, sensitivity analyses, and label-efficiency experiments assess the representation's components, parameter choices, and data efficiency.
\end{list}

\AtBeginEnvironment{equation}{%
  \setlength{\abovedisplayskip}{3pt plus 1pt minus 1pt}%
  \setlength{\belowdisplayskip}{2pt plus 1pt minus 1pt}%
  \setlength{\abovedisplayshortskip}{0.5pt plus 1pt minus 0.5pt}%
  \setlength{\belowdisplayshortskip}{2pt plus 1pt minus 1pt}}
\methodsection{Methodology}
We use the following notation throughout this section. Bold uppercase and lowercase symbols denote matrices and vectors, respectively.

Let $\D=\{(\mat{X}_i,y_i)\}_{i=1}^{N}$ contain $N$ labeled sequences $\mat{X}_i\in\R^{T_i\times d}$ with $T_i$ time points, $d$ channels, and labels $y_i\in\{1,\ldots,C\}$ for $C$ classes.

As shown in our framework in Figure~\ref{fig:framework}, \method maps training sequences and a query $\mat{X}_*$ to a common feature space.

\methodsubsection{Shared Coordinate System}
A common basis makes each operator coefficient refer to the same coordinate directions across examples. We interpolate missing channel values, normalize channels using training means and scales, and append first differences. A principal component analysis (PCA) projection~\cite{pca} gives
\begin{equation}
 \mat{Z}_i=\mathcal I_L\!\left(
 ([\widetilde{\mat{X}}_i,\Delta\widetilde{\mat{X}}_i]-\vect{\mu})\mat{P}\right)
 \in\R^{L\times r}.
 \label{eq:latent}
\end{equation}
Here $\widetilde{\mat{X}}_i$ is the normalized signal; $\Delta$ takes first differences with a zero initial row; brackets concatenate channels. Training data determine $\vect{\mu}\in\R^{2d}$ and $\mat{P}\in\R^{2d\times r}$; the center is subtracted from every row. Rank $r$ is capped by $r_0$ and available dimensions. The operator $\mathcal I_L$ interpolates at $L$ equally spaced normalized times; each state is a column vector $\vect{z}_t\in\R^r$.

\methodsubsection{Local Dynamics and Shared Regimes}
Local operators describe short intervals, while shared regimes summarize recurring forms of evolution. We extract windows of $m$ states at stride $s$ and fit
\begin{equation}
 \begin{aligned}
 (\mat{A}_w,\vect{b}_w)=\operatorname*{arg\,min}_{\mat{A},\vect{b}}\;&\frac{1}{m-1}
 \sum_{t\in\mathcal T_w}\lVert\vect{z}_{t+1}-\mat{A}\vect{z}_t-\vect{b}\rVert_2^2\\
 &+\lambda_w\lVert\mat{A}\rVert_F^2+\epsilon\lVert\vect{b}\rVert_2^2.
 \end{aligned}
 \label{eq:local}
\end{equation}
Here $\mathcal T_w$ indexes the $m-1$ adjacent pairs; $\mat{A}_w\in\R^{r\times r}$ and $\vect{b}_w\in\R^r$ define local affine evolution. The ridge penalty~\cite{ridge} adapts to input scale via $\lambda_w=\lambda\max\{\operatorname{tr}(\mat{G}_w)/(r+1),\epsilon\}$, where $\mat{G}_w$ averages outer products of $[\vect{z}_t^\top,1]^\top$ and $\epsilon>0$ is a numerical floor.

Training windows define a PCA encoding of the standardized concatenation of $\operatorname{vec}(\mat{A}_w-\mat{I}_r)$, $\vect{b}_w$, and normalized residuals; $\operatorname{vec}$ stacks matrix entries and $\mat{I}_r$ is the identity. Appending velocity summaries and standardizing yields $\vect{u}_w$. Mini-batch k-means fits $K$ shared centers $\vect{c}_k$, with soft assignments
\begin{equation}
 q_{wk}=\frac{\exp(-\lVert\vect{u}_w-\vect{c}_k\rVert_2/\tau)}
 {\sum_{a=1}^{K}\exp(-\lVert\vect{u}_w-\vect{c}_a\rVert_2/\tau)}.
 \label{eq:soft}
\end{equation}
Weight $q_{wk}$ measures how closely window $w$ resembles regime $k$. The temperature $\tau$ is the median training distance to the nearest center, with a positive floor. With $\vect{q}_w=(q_{w1},\ldots,q_{wK})^\top$ and $M$ windows, we retain weighted operator moments, occupancy, dwell times, and lagged transitions:
\begin{equation}
 \mat{P}^{(\ell)}=\frac{1}{M-\ell}\sum_{w=1}^{M-\ell}\vect{q}_w\vect{q}_{w+\ell}^{\top}.
 \label{eq:transitions}
\end{equation}
The matrix $\mat{P}^{(\ell)}\in\R^{K\times K}$ records joint regime weights at lag $\ell$, with $\sum_{a,b}P^{(\ell)}_{ab}=1$. Occupancy averages regime weights; dwell times measure consecutive dominant state runs.

\begin{table*}[!t]
 \centering
 \tighttablecaption
 \caption{Test accuracy (\%) on eight UEA datasets grouped by domain. Avg. equally weights datasets; Rank orders Avg. (1 = best). Best in bold; second best underlined.}
 \label{tab:results}
 \fontsize{9}{10.2}\selectfont
 \newcommand{\tableSD}[1]{{\fontsize{7.2}{8}\selectfont$\pm$\,#1}}
 \newcommand{\domainhead}[1]{{\fontsize{8.2}{9}\selectfont\bfseries #1}}
 \newcommand{\familyhead}[1]{\multicolumn{11}{@{}l}{\fontsize{8.2}{9}\selectfont\bfseries #1}\\}
 \newcommand{\tableSecond}[1]{\begingroup\setbox0=\hbox{#1}\leavevmode\rlap{\raisebox{-1pt}[0pt][0pt]{\rule{\wd0}{0.24pt}}}\box0\endgroup}
 \newcommand{\tableKeyRow}{\rule[-1.88pt]{0pt}{9.8pt}}
 \resizebox{\tableScale\textwidth}{!}{%
 \setlength{\tabcolsep}{4.2pt}%
 \setlength{\arrayrulewidth}{0.25pt}%
 \setlength{\aboverulesep}{1pt}%
 \setlength{\belowrulesep}{1pt}%
 \setbox\strutbox=\hbox{\vrule height7.6pt depth1.4pt width0pt}%
 \renewcommand{\arraystretch}{1}%
 \begin{tabular}{@{}l|cc|cc|cc|c|c|c|c@{}}
 \toprule[1.10pt]
 \raisebox{-0.85ex}{\textbf{Method}} & \multicolumn{2}{c|}{\domainhead{Cortical control}} & \multicolumn{2}{c|}{\domainhead{Motion}} & \multicolumn{2}{c|}{\domainhead{Speech}} & \domainhead{Heart sounds} & \domainhead{Spectroscopy} & \raisebox{-0.85ex}{\textbf{Avg. $\pm$ SD}} & \raisebox{-0.85ex}{\textbf{Rank}} \\
 \cmidrule(lr){2-3}\cmidrule(lr){4-5}\cmidrule(lr){6-7}\cmidrule(lr){8-8}\cmidrule(lr){9-9}
 & \textbf{SCP1} & \textbf{SCP2} & \textbf{HW} & \textbf{UW} & \textbf{JV} & \textbf{SAD} & \textbf{HB} & \textbf{EC} & & \\
 \midrule[0.55pt]
 \tableKeyRow{}Channels & 6 & 7 & 3 & 3 & 12 & 13 & 61 & 3 & -- & -- \\
 \midrule[0.40pt]
 \familyhead{Classical baselines}
 DTW~\cite{dtw} & 87.70 & 55.60 & 39.10 & 57.50 & 97.20 & 98.60 & \tableSecond{77.60} & 28.70 & 67.75\tableSD{0.44} & 14 \\
 XGBoost~\cite{xgboost} & 92.50 & 55.30 & 28.00 & 86.20 & 97.60 & 97.70 & 76.10 & 28.00 & 70.18\tableSD{0.25} & 11 \\
 \midrule[0.30pt]
 \familyhead{Neural sequence models}
 LSTM~\cite{lstm} & 71.80 & 50.80 & 6.90 & 64.70 & 92.00 & 12.30 & 73.20 & 28.00 & 49.96\tableSD{0.63} & 15 \\
 Informer~\cite{informer} & 90.40 & 54.40 & 32.00 & 86.50 & 97.00 & 98.70 & 76.90 & 30.20 & 70.76\tableSD{0.33} & 9 \\
 DLinear~\cite{dlinear} & 91.60 & 51.10 & 22.80 & 81.90 & 96.50 & 96.50 & 75.40 & 29.50 & 68.16\tableSD{0.26} & 12 \\
 InceptionTime~\cite{inceptiontime} & 80.61 & 51.22 & \textbf{59.11} & 87.25 & 98.16 & \tableSecond{99.48} & 75.12 & 27.76 & 72.34\tableSD{0.90} & 8 \\
 TimesNet~\cite{timesnet} & 85.67 & 51.11 & 27.74 & 85.56 & 95.14 & 98.88 & 72.59 & 25.48 & 67.77\tableSD{1.20} & 13 \\
 \midrule[0.30pt]
 \familyhead{Convolution + Ridge}
 MiniR + Ridge~\cite{minirocket} & 91.54 & 51.89 & 51.84 & 94.13 & \tableSecond{98.27} & 98.87 & 76.00 & 46.54 & 76.13\tableSD{0.47} & 6 \\
 MR + Ridge~\cite{multirocket} & \textbf{93.72} & 55.89 & 51.01 & \tableSecond{94.19} & 97.95 & 98.91 & 74.44 & 54.90 & 77.63\tableSD{0.47} & 4 \\
 HYDRA + MR + Ridge~\cite{hydra} & \tableSecond{93.65} & \tableSecond{56.67} & 51.46 & \tableSecond{94.19} & 98.05 & 98.97 & 76.20 & 54.52 & 77.96\tableSD{0.32} & 3 \\
 \midrule[0.30pt]
 \familyhead{Frozen tabular predictors}
 TS2TabPFN (v2.5)~\cite{ts2tabpfn} & 81.43 & 49.67 & 28.19 & 89.38 & 96.05 & 98.31 & 76.78 & 42.05 & 70.23\tableSD{0.18} & 10 \\
 MiniR + TabPFN~\cite{minirocket} & 91.13 & 53.44 & 50.87 & 92.50 & 97.95 & 98.84 & 77.37 & 45.02 & 75.89\tableSD{0.44} & 7 \\
 MR + TabPFN~\cite{multirocket} & 90.99 & 55.11 & 48.75 & 93.38 & 98.00 & 98.60 & 76.78 & 57.57 & 77.40\tableSD{0.43} & 5 \\
 MR + TabPFN (matched)~\cite{multirocket} & 88.46 & 51.78 & \tableSecond{53.72} & 92.00 & 97.57 & 98.34 & 76.39 & \tableSecond{74.45} & \tableSecond{79.09}\tableSD{0.40} & 2 \\
 \midrule[0.55pt]
 \tableKeyRow\textbf{SwitchPFN} & 91.40 & \textbf{57.56} & 49.72 & \textbf{95.44} & \textbf{99.08} & \textbf{99.52} & \textbf{78.24} & \textbf{90.04} & \textbf{82.63}\tableSD{0.45} & \textbf{1} \\
 \bottomrule[0.85pt]
 \end{tabular}}
\end{table*}

\methodsubsection{Class-Specific Dynamics Banks}
Class residuals measure which class best explains a sequence. A bank contains an affine model for each class $c$ and horizon $h\in\mathcal H$. For $\vect{\xi}_{it}=[\vect{z}_{it}^{\top},1]^\top$ and $t=1,\ldots,L-h$, each sequence contributes averaged moments:
\begin{equation}
 \mat{G}_{ih}=\frac{1}{L-h}\sum_t\vect{\xi}_{it}\vect{\xi}_{it}^{\top},\quad \mat{H}_{ih}=\frac{1}{L-h}\sum_t\vect{\xi}_{it}\vect{z}_{i,t+h}^{\top}.
 \label{eq:stats}
\end{equation}
Let $\mat{G}_{ch},\mat{H}_{ch}$ denote unnormalized class sums. For training row $i$, we subtract its own class contribution,
\begin{equation}
 (\mat{G}_{ch}^{(-i)},\mat{H}_{ch}^{(-i)})=(\mat{G}_{ch},\mat{H}_{ch})
 -\mathbf1[y_i=c](\mat{G}_{ih},\mat{H}_{ih}),
 \label{eq:loo}
\end{equation}
where $\mathbf1[y_i=c]$ indicates membership in class $c$. Counts and target second moments are adjusted too; the ridge fit divides sums by the remaining class count. An empty class uses a pooled model excluding $i$; query residuals use the complete training banks.

For prediction errors $\mat{E}_{ch}(\mat{X})$ and target states $\mat{Y}_h(\mat{X})$, we compute
\begin{equation}
 r_{ch}(\mat{X})=\log\!\left(1+
 \frac{\operatorname{RMS}(\mat{E}_{ch}(\mat{X}))}
 {\max\{\operatorname{RMS}(\mat{Y}_h(\mat{X})),\epsilon\}}\right).
 \label{eq:residual}
\end{equation}
RMS is the root mean square over entries. Five features per class capture the mean and spread of horizon scores, temporal residual variance, Gaussian negative log-likelihood, and the margin to the best competing class.

\methodsubsection{Feature Construction and Frozen Prediction}
The four candidate groups in Fig.~\ref{fig:framework} combine local switching, class residuals, paths, and signal summaries into $\vect{f}(\mat{X})$. Path features use truncated logsignatures of latent and velocity paths~\cite{signature,deepsignature}; generic summaries include moments, quantiles, spectral bands, autocorrelations, and cross-channel statistics.

\textbf{Hybrid selection} reserves $B_0=\operatorname{round}(\alpha B)$ columns in a fixed order, then selects the remaining columns by training label Fisher scores. Here $B$ is the final width, $\alpha$ is the reserved fraction, and $\operatorname{round}$ rounds to the nearest integer. Fisher scores compare between-class separation with within-class variation. For selected indices $J$, feature standardization gives
\begin{equation}
 \phi_j(\mat{X})=\frac{f_j(\mat{X})-\bar f_j}{\max(s_j,\epsilon)},\quad j\in J,
 \qquad |J|=B,
 \label{eq:table}
\end{equation}
using training mean $\bar f_j$ and standard deviation $s_j$. Stacking selected components gives $\vect{\phi}(\mat{X})\in\R^B$. Training rows form the context; query probabilities are
\begin{equation}
 \widehat{\vect{p}}_*=F_{\vect{\theta}_0}\!\left(
 \{(\vect{\phi}_i^{\mathrm{train}},y_i)\}_{i=1}^{N},\vect{\phi}^{\mathrm{query}}(\mat{X}_*)\right).
 \label{eq:predict}
\end{equation}
Here $\vect{\phi}_i^{\mathrm{train}}$ encodes $\mat{X}_i$ and $\widehat{\vect{p}}_*\in\R^C$ contains class probabilities. Weights $\vect{\theta}_0$ stay fixed; context and query rows use sequence-excluded and full class banks, respectively. Each validation fold fits all representation components on its training portion.

\section{Experiments}
\vspace{-7pt}
\subsection{Experimental Setup}\label{sec:setup}
\setlength{\baselineskip}{10.45pt plus .25pt}
\textbf{Datasets.} We evaluate eight University of East Anglia (UEA) tasks~\cite{uea2018} covering five signal domains: cortical control (SelfRegulationSCP1/2, SCP1/2), motion (Handwriting, HW; UWaveGestureLibrary, UW), speech (JapaneseVowels, JV; SpokenArabicDigits, SAD), heart sounds (Heartbeat, HB), and spectroscopy (EthanolConcentration, EC). The selection spans 3--61 channels, 2--26 classes, and varied sequence lengths. We use official UEA train/test splits for multivariate evaluation; UCR contains univariate series~\cite{ucr2019}. Selected UEA subsets are also used in prior multivariate classification work~\cite{li2023time}.

\textbf{Implementation.} We set $L=128$, $r_0=12$, $m=8$, $s=4$, $K=12$, and $\lambda=10^{-2}$. Horizons are $\mathcal H=\{1,2,4\}$; lags are $\{1,2\}$. Order-three logsignatures use up to four coordinates plus time in 1/2/4 segments. With $B=1{,}024$ and $\alpha=0.25$, 256 prefix columns are reserved and 768 ranked. Eight TabPFN-v3 estimators each receive at most 200 columns, jointly covering the table~\cite{tabpfn3}. Settings are frozen after training-only selection. Each test repeat fits on the full official training split and evaluates the official test split. Five repeats (seeds 2027--2031) equally average the eight dataset accuracies.

\textbf{Baselines.} We compare four families in Table~\ref{tab:results} and Fig.~\ref{fig:main}: classical baselines (DTW and XGBoost~\cite{dtw,xgboost}); neural sequence models (LSTM, Informer, DLinear, InceptionTime, and TimesNet~\cite{lstm,informer,dlinear,inceptiontime,timesnet}); convolution + Ridge (MiniRocket, MultiRocket, and HYDRA + MultiRocket~\cite{minirocket,multirocket,hydra}); and frozen tabular predictors (TabPFN with MiniRocket or MultiRocket, and TS2TabPFN~\cite{ts2tabpfn}). The matched MultiRocket + TabPFN control uses MultiRocket features with our $B=1{,}024$ budget and TabPFN-v3 predictor. DTW, XGBoost, LSTM, Informer, and DLinear use historical three-run summaries; other methods use five repeats. Informer and DLinear are classification adaptations; TS2TabPFN uses v2.5.

\subsection{Main Results}
\begin{figure}[H]
 \tightfigurecaption
 \centering\includegraphics[width=\figThreeScale\columnwidth,trim=0 5bp 0 0,clip]{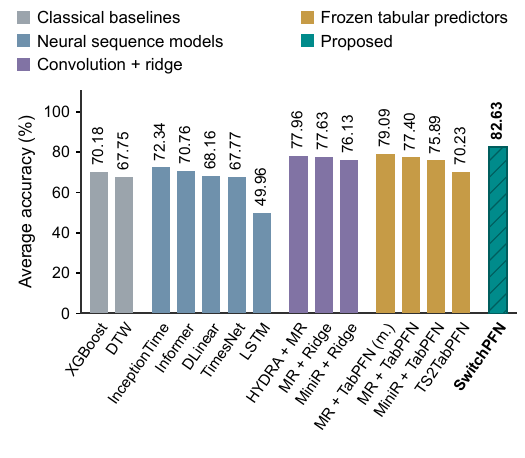}\par
 \caption{Mean test accuracy for the methods in Table~\ref{tab:results}. Colors group method families; hatching marks \method; (m.) denotes matched. HYDRA + MR uses Ridge, and TS2TabPFN is adapted v2.5.}
 \label{fig:main}
\end{figure}

In Table~\ref{tab:results} and Fig.~\ref{fig:main}, \method achieves the highest mean, \mbox{$82.63\pm0.45\%$}, versus \mbox{$79.09\pm0.40\%$} for matched MultiRocket + TabPFN: a gain of 3.54 percentage points (pp), or a 4.47\% relative improvement. Gains over HYDRA + MultiRocket Ridge and MultiRocket Ridge are 4.66 and 5.00 pp, respectively. Full MultiRocket + TabPFN reaches 77.40\%.

\raggedcolsend
\setlength{\baselineskip}{10.45pt plus .65pt}
\setlength{\intextsep}{5pt plus 1pt minus 1pt}
\vspace{0pt plus 1pt}

\method beats the matched control on seven of eight datasets, including at least one in each of the five domains. EC gives the largest gain (+15.59 pp); excluding EC, \method still wins six of seven datasets, with a mean gain of 1.81 pp. Overall, these gains span diverse multivariate signal domains.

\begin{figure}[H]
 \tightfigurecaption
 \centering\includegraphics[width=\figFourScale\columnwidth]{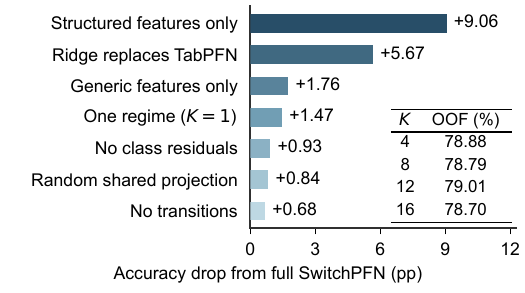}\par
 \caption{Full-minus-variant mean test accuracy (pp); positive values favor \method. Inset: training-set OOF validation accuracy (\%) versus $K$.}
 \label{fig:ablation}
\end{figure}

\methodsubsection{Model Analysis}
\textbf{Ablation study.} In Fig.~\ref{fig:ablation}, a single regime ($K=1$) loses 1.47 pp. Removing class residuals, using a random shared projection, and replacing TabPFN with Ridge lose 0.93, 0.84, and 5.67 pp, respectively. Structured-only and generic-only features lose 9.06 and 1.76 pp, supporting joint use. Removing transitions loses 0.68 pp. Together, these ablations support the contribution of each tested component.

\vspace{0pt plus 1pt}
\textbf{Sensitivity.} Table~\ref{tab:sensitivity} and the inset in Fig.~\ref{fig:ablation} use out-of-fold (OOF) validation within official training splits: three repeats (seeds 2042--2044), with five folds except two for HW. Each fold is predicted by a model fitted on the other folds. The frozen reference configuration scores 79.01\%; increasing $r_0$ to 24 gives 79.62\%, while retaining only the first 1,024 columns ($\alpha=1$) gives 69.34\%. Accuracy is more sensitive to feature selection than to rank or budget.\par

\begin{table}[H]
 \centering
 \tighttablecaption
 \caption{Training-only sensitivity: mean $\pm$ SD (\%) across three repeats of the eight-dataset OOF average. Bold settings identify the frozen recipe; its result is shared across sweeps.}
 \label{tab:sensitivity}
  \fontsize{9.1}{10.5}\selectfont
  \newcommand{\sensitivitySD}[1]{$\pm$\,#1}
  \setlength{\tabcolsep}{4.0pt}
 \setlength{\arrayrulewidth}{0.25pt}
 \setlength{\aboverulesep}{1.0pt}
 \setlength{\belowrulesep}{1.0pt}
 \renewcommand{\arraystretch}{1.08}
 \begin{tabular*}{\columnwidth}{@{\extracolsep{\fill}}cc|cc|cc@{}}
 \toprule[0.75pt]
 $r_0$ & Accuracy & $B$ & Accuracy & $\alpha$ & Accuracy \\
 \midrule[0.45pt]
 6 & 78.32\sensitivitySD{0.31} & 256 & 78.31\sensitivitySD{0.49} & 0 & 78.84\sensitivitySD{0.16} \\
 \textbf{12} & 79.01\sensitivitySD{0.59} & 512 & 79.16\sensitivitySD{0.34} & \textbf{0.25} & 79.01\sensitivitySD{0.59} \\
 24 & 79.62\sensitivitySD{0.67} & \textbf{1024} & 79.01\sensitivitySD{0.59} & 0.5 & 79.00\sensitivitySD{0.39} \\
 -- & -- & 1536 & 78.73\sensitivitySD{0.35} & 1 & 69.34\sensitivitySD{1.03} \\
 \bottomrule[0.75pt]
 \end{tabular*}
\end{table}
\unskip

\begin{figure}[H]
 \tightfigurecaption
 \centering\includegraphics[width=\figFiveScale\columnwidth]{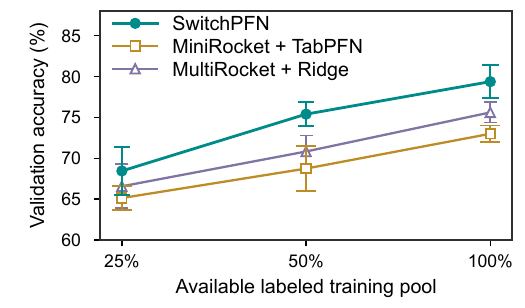}\par
 \caption{Training-only label efficiency: mean $\pm$ sample SD over three holdouts. Pool fractions exclude the validation samples.}
 \label{fig:labels}
\end{figure}

\vspace{0pt plus 1pt}
\textbf{Data efficiency.} Figure~\ref{fig:labels} uses fixed validation holdouts within official training splits (seeds 2045--2047). With 25\%, 50\%, and 100\% of the remaining pool, \method reaches validation accuracies of 68.45\%, 75.40\%, and 79.38\%, respectively. Gains over MiniRocket + TabPFN are 3.32, 6.66, and 6.38 pp; \method also exceeds MultiRocket + Ridge at each fraction. Representation fitting averages 66.72 s on shared RTX 4090 hardware; the full SAD pipeline takes 560.63 s. The advantage persists even with a quarter of the labeled training pool.

\vspace{0pt plus 1pt}

\vspace{-12pt plus 1pt}
\methodsection{Conclusion and Future Work}
\vspace{-4pt}
We presented \method, which represents local dynamics and regime transitions in a shared feature space for classification with a frozen TabPFN predictor. Shared coordinates, a common regime codebook, and class-specific residual features provide complementary information. Experiments on eight UEA datasets demonstrate improved mean accuracy and effective performance with limited labels. Future work will extend evaluation to more domains, refine feature selection through independent validation, and reduce the computational cost of feature construction.

\unnumsection{Acknowledgments}
The authors declare no conflicts of interest.

\vspace{0pt plus 1pt}
\unnumsection{Declaration of AI Use}
ChatGPT (OpenAI) was used solely for language editing to improve clarity and conciseness. All suggested revisions were reviewed and approved by the authors.
\vspace{0pt plus 1pt}
\unnumsection{Compliance with Ethical Standards}
This work uses only publicly available benchmark datasets and involves no collection of new data from human participants.
\label{body:end}
\par\penalty-10000
\raggedcolsend
\clearpage
\flushcolsend
\flushend
\linespread{.91}\selectfont
\apptocmd{\thebibliography}{\setlength{\itemsep}{0pt}}{}{}
\bibliographystyle{IEEEbib}

\end{document}